\documentclass{article}

\usepackage{arxiv}

\usepackage[utf8]{inputenc} % allow utf-8 input
\usepackage[T1]{fontenc}    % use 8-bit T1 fonts
\usepackage{hyperref}       % hyperlinks
\usepackage{url}            % simple URL typesetting
\usepackage{booktabs}       % professional-quality tables
\usepackage{amsmath}        % math environments
\usepackage{amsfonts}       % blackboard math symbols
\usepackage{nicefrac}       % compact symbols for 1/2, etc.
\usepackage{microtype}      % microtypography
\usepackage{graphicx}       % for including images
\usepackage{multirow}       % for multi-row table cells
\usepackage{makecell}       % for multi-line table cells
\graphicspath{ {./Images/} }

\title{Stack Transformer Based Spatial-Temporal Attention Model for Dynamic Sign Language and Fingerspelling Recognition}

\author{
  Koki Hirooka \\
  School of Computer Science and Engineering\\
  The University of Aizu\\
  Aizuwakamatsu 965-8580, Japan\\
  \And
  Abu Saleh Musa Miah \\
  School of Computer Science and Engineering\\
  The University of Aizu\\
  Aizuwakamatsu 965-8580, Japan\\
  \And
  Tatsuya Murakami \\
  School of Computer Science and Engineering\\
  The University of Aizu\\
  Aizuwakamatsu 965-8580, Japan\\
  \And
  Md. Al Mehedi Hasan \\
  Department of Computer Science and Engineering\\
  Rajshahi University of Engineering and Technology\\
  Rajshahi-6204, Bangladesh\\
  \And
  Yong Seok Hwang \\
  Department of Electronics Engineering\\
  Kwangwoon University\\
  Seoul 01897, South Korea\\
  \texttt{thestone@kw.ac.kr} \\
  \And
  Jungpil Shin\thanks{Corresponding author.} \\
  School of Computer Science and Engineering\\
  The University of Aizu\\
  Aizuwakamatsu 965-8580, Japan\\
  \texttt{jpshin@u-aizu.ac.jp} \\
}

\begin{document}
\maketitle

\begin{abstract}
Hand gesture-based Sign Language Recognition (SLR) serves as a crucial communication bridge between deaf and non-deaf individuals.
While Graph Convolutional Networks (GCNs) are common, they are limited by their reliance on fixed skeletal graphs.
To overcome this, we propose the Stacked Spatio-Temporal Attention Network (SSTAN), a novel Transformer-based architecture.
Our model employs a hierarchical, stacked design that sequentially integrates Spatial Multi-Head Attention (MHA) to capture intra-frame joint relationships and Temporal MHA to model long-range inter-frame dependencies.
This approach allows the model to efficiently learn complex spatio-temporal patterns without predefined graph structures.
We validated our model through extensive experiments on diverse, large-scale datasets (WLASL, JSL, and KSL).
A key finding is that our model, trained entirely from scratch, achieves state-of-the-art (SOTA) performance in the challenging fingerspelling categories (JSL and KSL).
Furthermore, it establishes a new SOTA for skeleton-only methods on WLASL, outperforming several approaches that rely on complex self-supervised pre-training.
These results demonstrate our model's high data efficiency and its effectiveness in capturing the intricate dynamics of sign language.
The official implementation is available at our GitHub repository: \href{https://github.com/K-Hirooka-Aizu/skeleton-slr-transformer}{https://github.com/K-Hirooka-Aizu/skeleton-slr-transformer}.
\end{abstract}


\section{Introduction}
\label{sec1}
According to the World Health Organization (WHO)\cite{who_hearing_loss_2024}, over 430 million people worldwide—more than 5\% of the global population—require rehabilitation to address disabling hearing loss, and this number is expected to rise to over 700 million by 2050.
Automatic Sign Language Recognition (SLR) has therefore emerged as a socially impactful research field dedicated to bridging the communication gap between Deaf and hearing individuals.
Sign language (SL), the primary communication medium for the Deaf community, is a visual-spatial language characterized by its own grammar and lexicon.
It conveys meaning through a combination of manual components—such as hand shapes, body postures, and movements—and non-manual elements like facial expressions and head motions.
Core tasks in Sign Language Understanding (SLU) include Isolated Sign Language Recognition (ISLR) for word-level recognition, Continuous Sign Language Recognition (CSLR) for recognizing continuous sign sequences, and Sign Language Translation (SLT) for generating spoken-language translations.

Currently, due to the rapid advancement of deep learning, various automatic sign language recognition systems are being developed.
The data modalities used in sign language research can be broadly categorized into vision-based and sensor-based modalities.
Sensor-based modalities capture human motion directly using wearable devices attached to the body. This approach is generally less affected by environmental factors and often achieves higher accuracy compared to vision-based methods. However, due to the inconvenience of wearing sensors and the high cost of the equipment, sensor-based approaches are not well-suited for practical real-time applications.
In contrast, vision-based modalities collect data using visual sensors such as cameras. This approach offers advantages in terms of lower cost and greater suitability for real-time applications.
Furthermore, various models have been proposed within the vision-based modality, depending on the type of input data such as RGB, depth, and skeleton.
Specificlly, skeleton-based data is typically modeled using graph neural networks (GNNs) or graph convolutional networks (GCNs).
These are well-suited for capturing the spatial and structural relationships between joints in the body, and the development of skeleton-based human action recognition and sign language recognition models, spearheaded by Spatial Temporal Graph Convolutional Networks (ST-GCN)\cite{yan2018spatial}, is advancing rapidly.
Li et al.~\cite{li2020word} have proposed different models for sign language recognition tasks depending on the data modality.
For RGB data, they introduced approaches that combine 2D Convolutional Neural Networks (2D CNNs) with Gated Recurrent Units (GRUs)\cite{cho2014learning}, as well as methods based on Inflated 3D ConvNets (I3D)\cite{carreira2017quo}.
For skeleton data, they also proposed models employing GRUs and pose-based temporal graph convolution networks (Pose-TGCN) to effectively capture temporal dynamics and spatial relationships among joints.
Liu et al.\cite{SKIM} have proposed a channel-wise Graph Neural Network (GNN) that uses Channel-wise Topology Refinement Graph Convolution (CTR-GC)\cite{chen2021channel} as its core backbone.
This network has multi-scale recognition capability and performs temporal reweighting for each keypoint.
Furthermore, the Hu et al.\cite{SignBERT, SignBERT+} have proposed an integrated model that employs a Graph Convolution Network for gesture state embedding and spatial embedding, and a BERT\cite{devlin2019bert} to capture temporal context.

The reason GCN continues to be adopted lies in its ability to incorporate the human body's structural prior knowledge as an ``inductive bias'' into the graph structure model, enabling efficient learning.
However, these models inherently possess limitations stemming from their reliance on fixed structures, which form the foundation of their success. The physical skeletal graphs used in GCNs struggle to directly capture the relationships between joints that are physically distant yet semantically strongly connected, such as the hands and head during actions like ``clapping'' or ``making a phone call.'' Consequently, their ability to model coordinated whole-body movements and context-dependent dynamic interactions between joints is limited, contributing to a plateau in recognition accuracy.

To overcome the limitations of fixed graph structures inherent in GCNs, this study focuses on Transformers\cite{vaswani2017attention}, which do not require predefined graph structures.
The self-attention mechanism, the core of the Transformer, dynamically computes relationships among all joints based on the task's context, effectively modeling global, coordinated movements and long-range temporal dependencies.

Therefore, we propose Stacked Spatio-Temporal Attention Network (SSTAN), a novel Transformer-based architecture for sign language and fingerspelling recognition.
Our model is built upon a post-normalization framework and is specifically designed to sequentially integrate spatial (intra-frame) and temporal (inter-frame) attention, aiming to achieve high performance across diverse sign language datasets.

The primary contributions of this study are as follows:
\begin{enumerate}
\item We propose a novel Transformer architecture, Stacked Spatio-Temporal Attention Network (SSTAN), which employs a post-normalization framework to sequentially integrate spatial and temporal dependencies using multi-head self-attention.
\item We demonstrate that our model achieves state-of-the-art performance on the Japanese and Korean fingerspelling datasets, and highly competitive results on the large-scale WLASL dataset, all without the need for pre-training.
\end{enumerate}
These contributions establish an effective and robust method for Sign Language Recognition (SLR), advancing the state-of-the-art by efficiently capturing complex spatio-temporal dynamics.

The presented work is organized as follows: Section \ref{sec2} summarizes the existing research work and problems related to the presented work, Section \ref{sec3} describes the benchmark and proposed Korean sign language datasets, and Section \ref{sec4} describes the architecture of the proposed system—Section \ref{sec5} demonstrated the evaluation performed with state of the art comparison.
In Section~\ref{sec6}, draw the conclusion and future work.

\section{Related Work}
\label{sec2}
\subsection{Sign Language and Fingerspelling Recognition}
As mentioned in the previous subsection, SLR is a specialized and challenging sub-domain of HAR.
Sign language recognition tasks are generally categorized into three main types: ISLR, CSLR, and SLT.
ISLR focuses on recognizing individual lexical signs, while CSLR and SLT aim to convert sign sequences into word sequences or natural language sentences.
Although ISLR deals with isolated signs, it remains an essential research direction, as it often serves as a foundation for more complex tasks such as CSLR and SLT\cite{zuo2024towards}.
As this study specifically focuses on the ISLR task, we review prior work on fingerspelling and isolated SLR separately, organizing them by their methodological evolution.

\subsubsection{Fingerspelling Recognition}
Fingerspelling recognition, which involves identifying the alphabet using handshapes, serves as a foundational task in SLR.
Early approaches primarily relied on hand-crafted features and traditional classifiers.
For instance, Pugeault and Bowden~\cite{6130290} utilized Gabor filters with Random Forests, while others employed descriptors like SIFT combined with SVMs~\cite{noble2006support, 1053964, estrela2013sign, 7033173, rioux2014sign}.

With the advent of deep learning, Convolutional Neural Networks (CNNs) became dominant for processing image data.
Ameen and Vadera~\cite{ameen2017convolutional} applied ConvNets~\cite{lecun1998convolutional} to depth and intensity images. Transfer learning with models like VGG19~\cite{crespo2019gesture, simonyan2014very} and InceptionV3~\cite{das2018sign, szegedy2016rethinking} further pushed accuracies to over 90\%.
Recently, Miah et al.~\cite{app12083933} demonstrated the efficacy of custom CNN architectures on various datasets~\cite{rafi2019image, Shorif2021KUBdSL, islam2018ishara}.

Parallel to vision-based methods, skeleton-based approaches have gained traction due to their robustness to background noise.
Shin et al.~\cite{s21175856} utilized keypoints extracted by MediaPipe~\cite{Zhang2020MediaPipeHO}, classifying features like joint angles with SVMs and LightGBM~\cite{ke2017lightgbm}.
More recently, hybrid approaches combining hand-crafted and CNN-extracted features from skeleton data have shown promise~\cite{10529244}.
While these methods achieve high accuracy for static signs, recognizing dynamic fingerspelling in continuous streams remains a challenge that necessitates advanced temporal modeling beyond simple frame-wise classification.

\subsubsection{Isolated Sign Language Recognition}
Moving beyond character-level recognition, word-level SLR introduces significant complexity due to the involvement of holistic upper-body movements and dynamic temporal evolution.

\paragraph{Early Deep Learning Approaches (CNN \& RNN)}
Initial deep learning attempts leveraged 3D-CNNs~\cite{7177428} to capture spatio-temporal features from multi-channel inputs (RGB, depth, skeleton). Similarly, architectures like I3D~\cite{carreira2017quo} proved effective for large-scale datasets like MS-ASL~\cite{joze2018ms}.
To model sequential dependencies, Recurrent Neural Networks (RNNs), including GRUs and LSTMs, were widely adopted~\cite{li2020word}.
However, while effective temporally, simple RNN-based models often struggle to explicitly capture the complex spatial structural relationships between body joints.

\paragraph{Graph Convolutional Networks (GCNs)}
To address the spatial limitations of RNNs, GCN-based methods became a standard for skeleton-based SLR.
Pose-TGCN~\cite{li2020word} demonstrated that modeling the body as a graph improves accuracy over RNNs.
Subsequent works like Miah et al.~\cite{electronics12132841} introduced multi-stream GCNs to incorporate bone and motion information.
Liu et al.\cite{SKIM} further advanced this by using CTR-GCN~\cite{chen2021channel} to dynamically refine the graph topology, tackling the issue of imbalanced keypoint density with their PANAM module.
Patra et al.~\cite{patra2024hierarchical} proposed Hierarchical Windowed Graph Attention to focus on meaningful subgraphs efficiently.
Despite their success, these GCN methods fundamentally rely on predefined or learned graph structures. This reliance can limit their ability to capture long-range dependencies between physically distant but semantically related joints (e.g., coordinated movements of the left and right hands) that are not directly connected in the skeletal graph.

\paragraph{Self-Supervised Learning \& Transformers}
Most recently, to overcome data scarcity and capture global context, Self-Supervised Learning (SSL) and Transformers have been introduced.
Inspired by BERT~\cite{devlin2019bert}, methods like SignBERT~\cite{SignBERT}, SignBERT+~\cite{SignBERT+}, and BEST~\cite{zhao2023BEST} pre-train on large-scale data to learn context-aware representations, achieving state-of-the-art performance.
Additionally, Zuo et al.~\cite{zuo2023natural} utilized soft labels based on semantic similarity, and Li et al.~\cite{scale88uni} integrated Large Language Models (LLMs) to bridge the gap between recognition and translation.
However, many of these high-performing models require computationally expensive pre-training on massive external datasets or rely on multi-modal inputs (RGB + Skeleton) to achieve top results.

\paragraph{Research Gap \& Proposed Approach}
As previously discussed, while GCNs are powerful for local spatial modeling, they are constrained by their graph structure, potentially missing crucial non-local interactions. Conversely, while Transformers excel at capturing long-range global context, applying them effectively to spatio-temporal skeleton data without heavy pre-training remains a challenge.
In this paper, we propose a fully Transformer-based approach to address these limitations.
By leveraging the multi-head self-attention mechanism for both spatial (intra-frame) and temporal (inter-frame) modeling, our method dynamically captures long-range dependencies without the constraints of a fixed graph.
Crucially, our approach establishes a new state-of-the-art for skeleton-only recognition trained entirely from scratch, demonstrating superior data efficiency compared to methods relying on complex self-supervised pre-training.
\section{Datasets}\label{sec3}
There are many sign language datasets publicly available for individual sign languages.
This study considered three datasets for ASL, JSL, and KSL.
The objective of this study is to demonstrate that the proposed method is not over-optimized for a single task or language. To this end, we intentionally selected datasets with different characteristics. Specifically, we included tasks of different granularities: (1) word-level sign language (WLASL) and (2) character-level fingerspelling (JSL/KSL). Furthermore, by using data from (3) different linguistic and cultural backgrounds (American, Japanese, and Korean), we also evaluate the cultural robustness of our method.
Among them, WLASL is a large-scale dataset for ASL.
Furthermore, to demonstrate the generality of the proposed method, we also evaluated it on two finger spelling datasets.

\begin{table}[htp]
\centering
\caption{Evaluated Dataset Description in the Study}
\label{Tab:Dataset_Table}
% l = left-aligned, c = center-aligned
\begin{tabular}{llccccc}
\toprule
\makecell{Dataset \\ Name} & Lang. & Signs & Sub. & \makecell{Total \\ Videos} & \makecell{Videos \\ Per Sign} & \makecell{Joints \\ Per Frame} \\
\midrule
WLASL & ASL & 2000 & 119 & 21089 & 10.5 & 67 \\
JSL & Japanese fingerspelling & 46 & 15 & --- & --- & 21 \\
KSL & Korean fingerspelling & 32 & 20 & --- & --- & 21 \\
\bottomrule
\end{tabular}
\end{table}

%%%%%%%%%%%%%%%%%%%%%%%%%%%%%%%%%%
\subsection{WLASL Dataset}\label{subsec3.1}
The WLASL dataset is one of the largest video resources for word-level American Sign Language (ASL) recognition, containing 68,129 videos featuring 20,863 unique ASL glosses collected from 20 diverse websites~\cite{li2020word}.
Each video captures a signer performing a single sign, typically from a frontal view, with varying background complexities.
This dataset aims to advance research in sign language recognition and improve communication between deaf and hearing communities.
To maintain consistency, gloss annotations with more than two English words were excluded, and glosses with fewer than seven video samples were removed to ensure adequate data for training and testing.
After preprocessing, the dataset comprises 34,404 videos for 3,126 glosses, organized by sample count.
WLASL is divided into four subsets based on vocabulary size: WLASL100, WLASL300, WLASL1000, and WLASL2000, representing the top 100, 300, 1,000, and 2,000 glosses, respectively.
These subsets enable the evaluation of word-level recognition challenges across different vocabulary sizes.The dataset also supports scalability testing for sign recognition methods.
WLASL remains an essential resource for developing innovative, large-scale sign language recognition systems.

\subsection{Japanese Sign Language (fingerspelling) Dataset}
Japanese Sign Language (JSL) datasets are limited, particularly for dynamic signs.
A new JSL dataset was created using an RGB camera to address this gap, capturing both static and dynamic signs.
This dataset comprises (76 kinds of japanese fingerspelling $\times$ 15 signers $=$) 1140 videos collected from 15 non-native signer aged 18 to 25.
Each participant recorded one video for each of the 76 JSL alphabet signs (46 static and 30 dynamic), ensuring diverse representation.
Informed consent was obtained from all subjects involved in the study.
The data collection process followed a structured approach.
Participants were seated before the camera and instructed to observe a finger-spelling model before mimicking the signs.
Recording one character took approximately 1 seconds, with each participant spending about 10 minutes to complete all 76 signs. Static signs have consistent data representation, while dynamic signs show variability in time and data volume, depending on individual performance.
The dataset provides valuable resources for studying static and dynamic JSL recognition, addressing a key challenge in developing comprehensive sign language models.
It offers detailed insights into the variability of dynamic gestures while maintaining consistency for static gestures.
Figures and tables associated with the dataset offer further context, showcasing examples of signs and highlighting the differences between static and dynamic gestures.
This dataset is a significant step toward advancing JSL recognition research.

\subsection{Korean Sign Language (fingerspelling) dataset}
We also used the Korean Sign Language (fingerspelling) dataset was collected in previous our lab's work, and it has 31 kinds of fingerspelling gestures.
This dataset has 620 videos were recorded by 20 non-native signers.Each video's duration is approximately 3 seconds.
In this experiment, we extracted 21 hands keypoints by Mediapipe from each frame.

\begin{figure}[htp]
    \centering
    \includegraphics[width=0.3\textwidth]{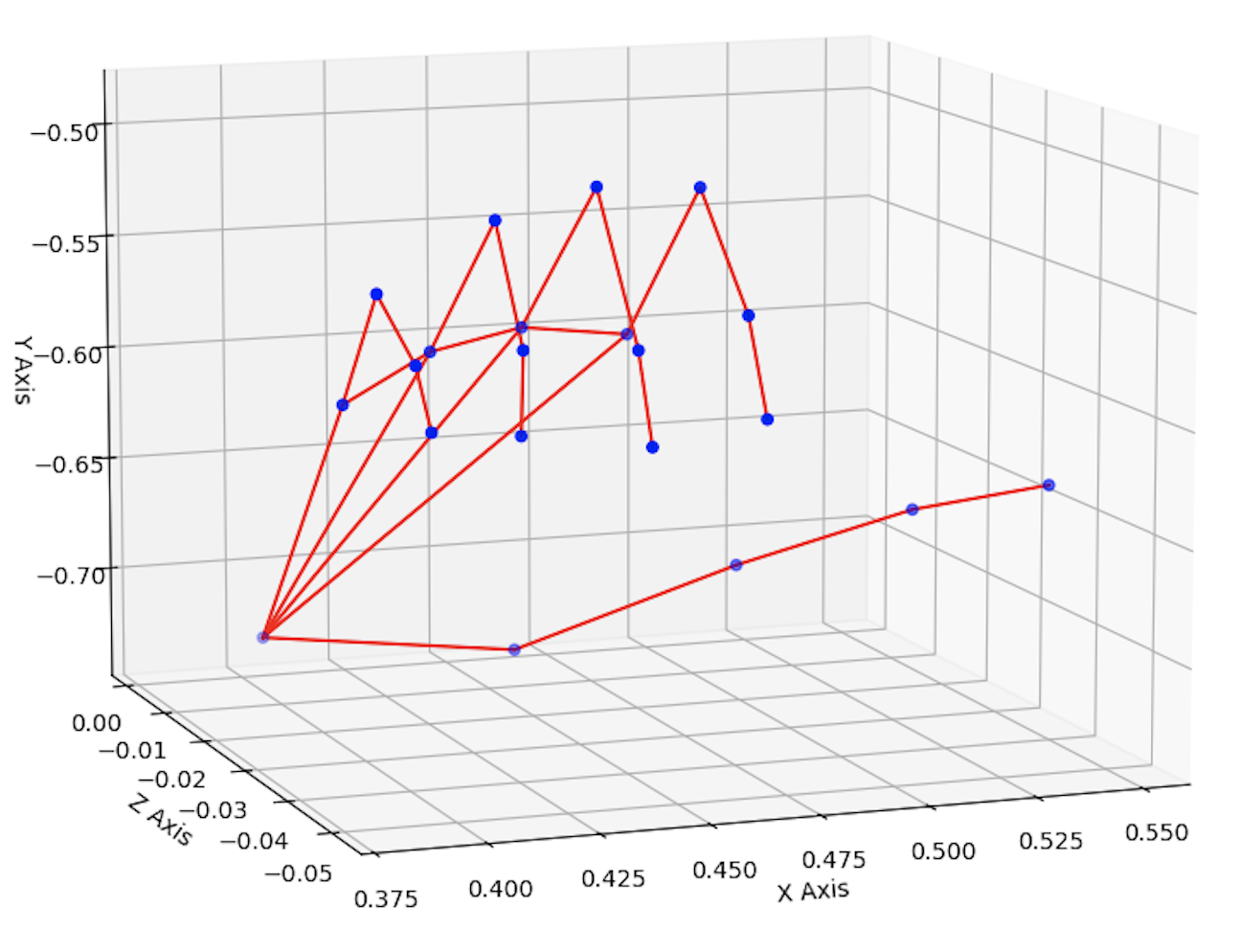}
    \caption{Skeleton graph for FignerSpelling.}
    \label{Figure:hand_skeleton}
\end{figure}

Figure\ref{Figure:hand_skeleton} shows the skeleton structure for JSL and KSL as input data.

\section{Proposed Methodology} \label{sec4}
\begin{figure}[htp]
    \centering
    \includegraphics[width=0.9\textwidth]{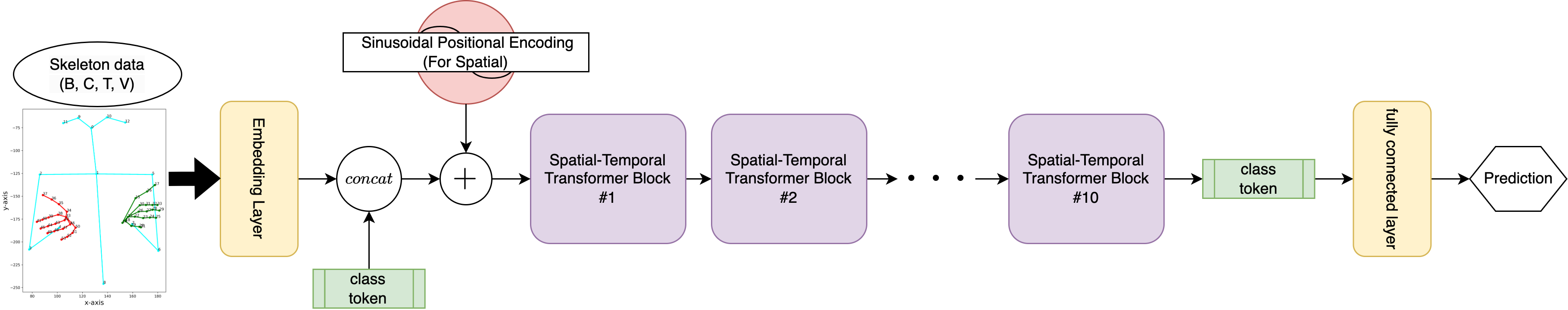}
    \caption{The overview of proposed architecture.}
    \label{Figure:5_Main_fig}
\end{figure}

Figure \ref{Figure:5_Main_fig} represents the structure of the proposed architecture based on the transformer.
Transformers have been used as deep learning methods for a variety of tasks (e.g., \textit{Natural Language Processing}, and \textit{Computer Vision}), and have demonstrated great performance and their effectiveness.
One of advantages in Transformer is allowing to capture the relationship between long-term in early stage.Another approach used in skeleton-based human action recognition and hand gesture recognition is the graph-convolution network, which uses a graph to convolve adjacent nodes.
This approach can capture the relationship between adjacent nodes, but it requires multiple convolution layers to capture the relationship between non-adjacent nodes.
For these reasons, we employed the Transformer in our approach.

\begin{figure}[htp]
    \centering
    \includegraphics[width=0.40\textwidth]{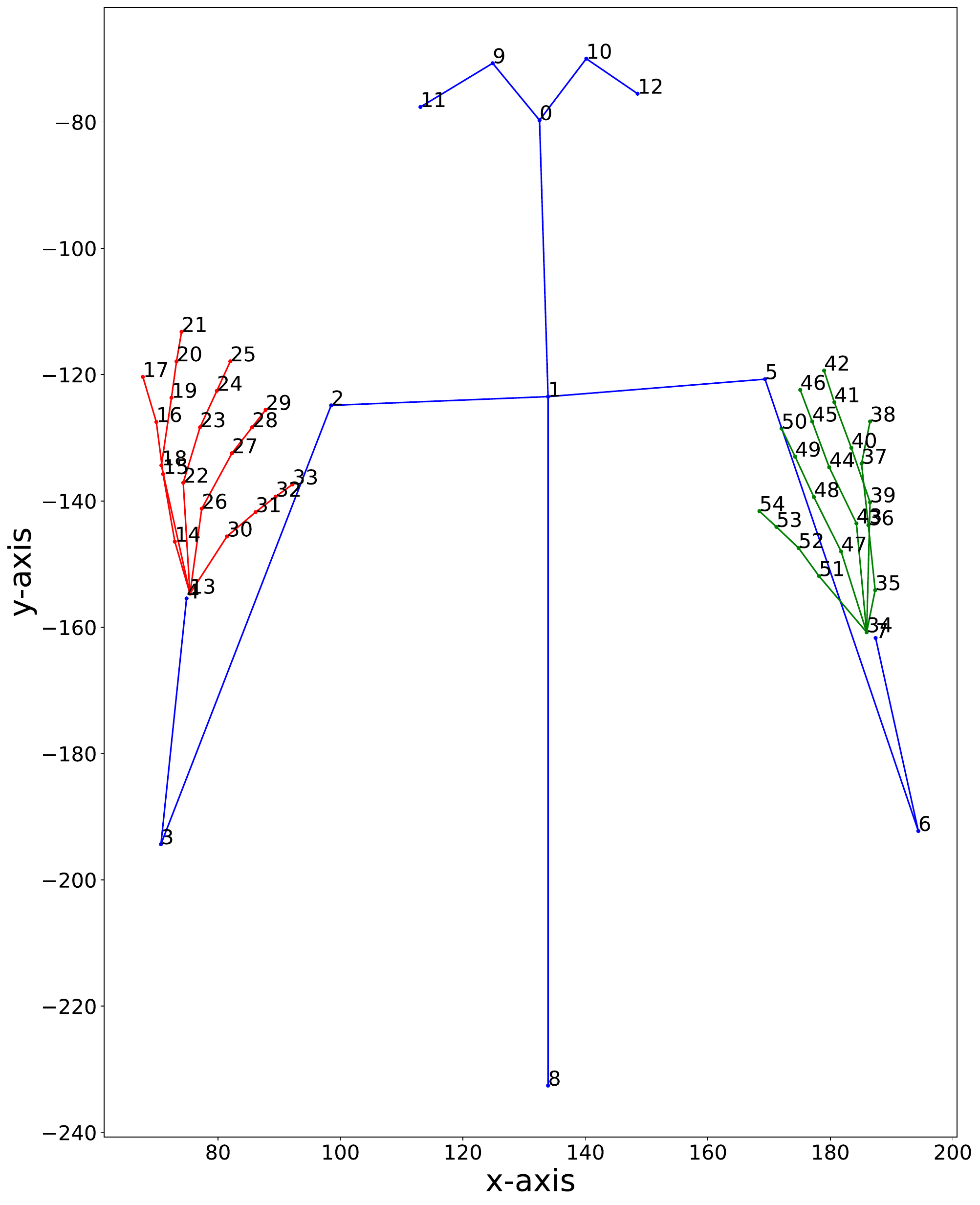}
    \caption{Pose and graph construction for WLASL.}
    \label{Figure:5_67keypoints}
\end{figure}

This diagram \ref{Figure:5_67keypoints} illustrates the structure of the skeleton input into our proposed method. Since our method is composed of Transformers, there is no need to predefine the skeleton structure as a graph, as is required with GCNs.

The skeleton data as input data fed into embedding layer which changes coordinates into embedding vectors.
Then, the embedding vectors were concatenated with \textit{class} token, added with Sinusoidal Positional Encoding, and passed multiple spatial temporal transformer blocks.
In the transformer block, embedding data was applied spatial self-attention, temporal self-attention, and feed forward network. After that, the \textit{class} token was extracted and fed into the fully connected layer.
Finally, we obtained the predicted label.
In the following subsections, each component of the proposed method is explained.

\subsection{Spatial Temporal Transformer Block}
\begin{figure}[htp]
    \centering
    \includegraphics[width=0.25\textwidth]{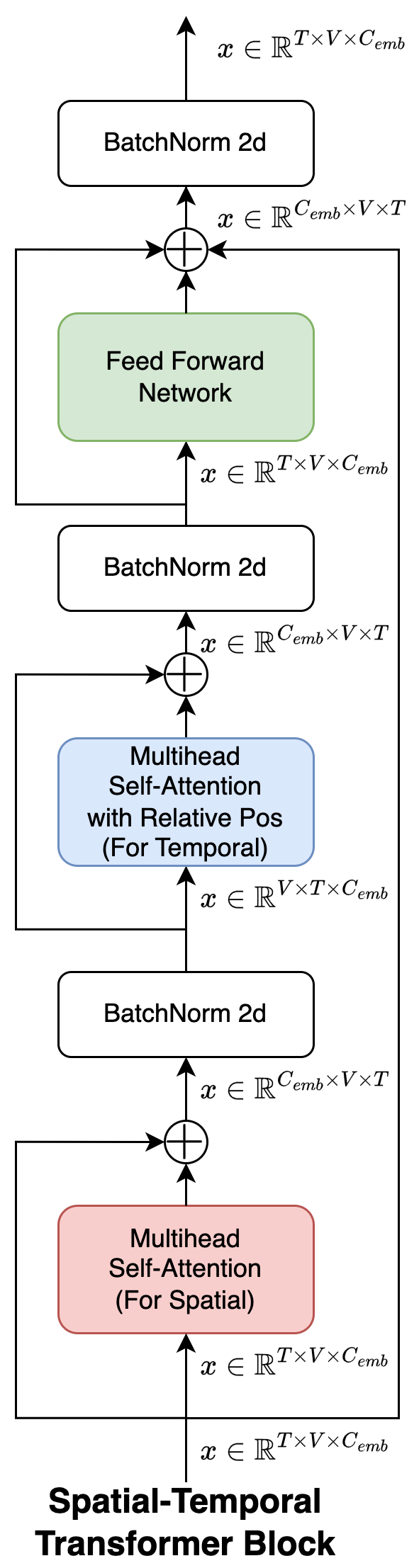}
    \caption{The internal structure of Spatial Temporal Transformer Block.}
    \label{Figure:STtransformer_block}
\end{figure}

Figure \ref{Figure:STtransformer_block} shows the internal structure of Spatial Temporal Transformer Block.
The design of this block was inspired by previous work in the field of action recognition \cite{yan2018spatial}, which demonstrated the effectiveness of a sequential spatial-then-temporal architecture. We adopted this design, positing that sequentially processing spatial and temporal features would be similarly effective for the sign language recognition task.
This block mainly consisted of 3 components : Multihead Self-Attention for spatial dimension, Multihead Self-Attention for tempral dimension with Relative Positional Encoding, and Feed Forward Network.
The position of the normalization layer in the Transformer is often discussed, and there are two types: Post-Norm, which places the normalization layer between residual blocks, and Pre-Norm, which places it inside the residual connection and places it before Self-attention.In our architecture, we adopted the Post-Norm structure because Post-Norm performs better for relatively shallow transformers\cite{xiong2020layer}, and we also observed that Post-Norm performs better than Pre-Norm in this experiment.
However, when Post-Norm is multi-layered, learning tends to become unstable.
Recently, a method called B2T Connection\cite{B2TConnection}, which skips all sub-layers, has been proposed to solve this problem, and it has been successful in multi-layering Post-Norm.
So, we used this technique for stabilizing the model learning.

\subsection{Multihead Self-Attention For Spatial dimension}
\begin{figure}[htp]
    \centering
    \includegraphics[width=0.4\textwidth]{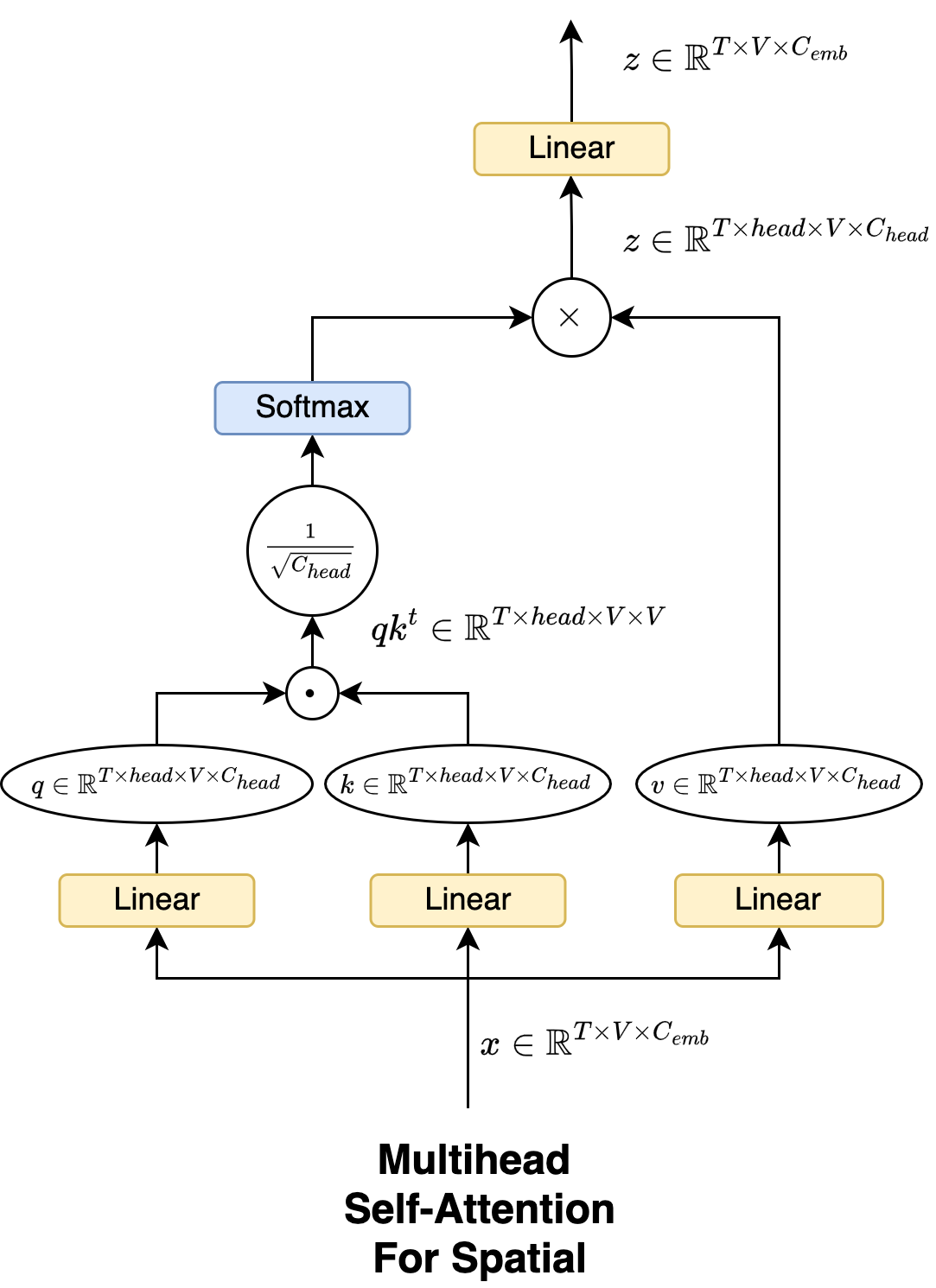}
    \caption{Multihead Self Attention for spatial dimension.}
    \label{Figure:spatial_mhsa}
\end{figure}

The input embedding vectors first met multihead self attention for spatial dimension.
Figure \ref{Figure:spatial_mhsa} shows the structure of multihead self attention for spatial dimension.
Let denote the input embedding vectors as $x\in\mathbb{R}^{T \times V \times C_{emb}}$.$T$ represents the length of the sequence, $V$ represents the number of vertices or nodes, and $C_{emb}$ represents the number of embedding channels.
Input embedding vectors fed into each linear layer to product \textit{query}, \textit{key}, and \textit{value}.
These process are expressed as follows:
\begin{equation}
  query = W_{q}(x)
\end{equation}
\begin{equation}
  key = W_{k}(x)
\end{equation}
\begin{equation}
  value = W_{v}(x)
\end{equation}
Then, the attention map was calculated by the dot product operation with \textit{query} and \textit{key}.
Finally, the output of attention was computed by $matmul$ operator with attention map and \textit{value}, and projected by output's linear layer as follows:
\begin{equation}
  output = W_{o}(MatMul(attention, v))
\end{equation}

As mentioned above, Graph Convolution can generally capture the relationships between adjacent nodes, but cannot capture the relationships between non-adjacent nodes.
On the other hand, by using Self-Attention, each node can access all nodes, and it plays a role in capturing the relationships between nodes at an early stage.

\subsection{Multihead Self-Attention with Relative Positonal Encoding for Temporal dimension}
\begin{figure}[htp]
    \centering
    \includegraphics[width=0.4\textwidth]{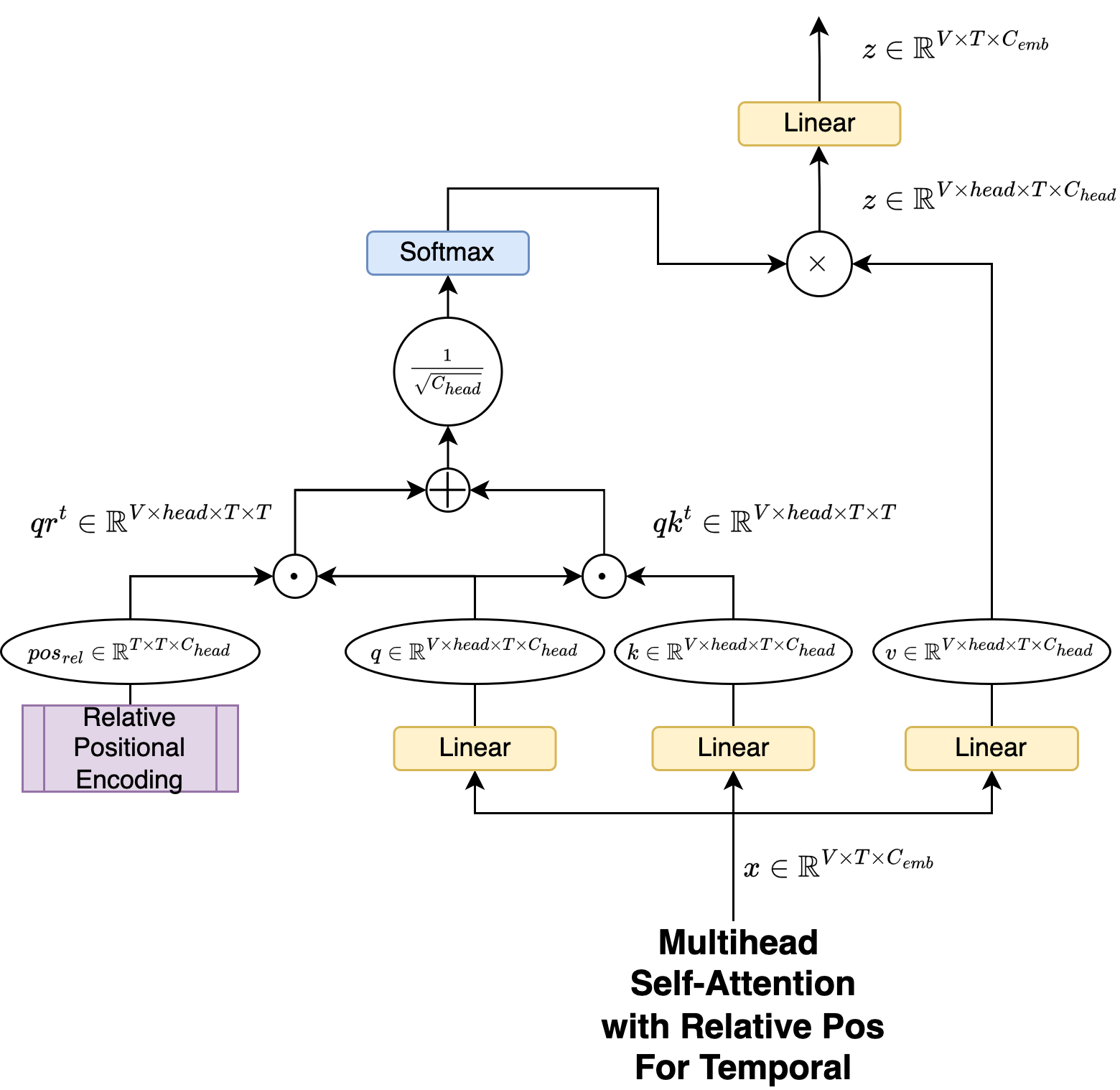}
    \caption{Multihead Self Attention with Relative Positional Encoding for temporal dimension.}
    \label{Figure:temporal_mhsa}
\end{figure}

Figure \ref{Figure:temporal_mhsa} shows the structure of Multihead Self Attention with Relative Positional Encoding.
This component aims to capture the relationships between different frames along a temporal dimension.However, this component does use relative position encoding along the temporal dimension.
This module is inspired by~\cite{wu2021rethinking}, and gets a temporal translation invariance that maintains robust performance even if a particular frame is moved forward or backward.
In our model, the input embedding vector was transposed in spatial and temporal dimensions before being passed to this module.

\subsection{Feed Forward Network}
Following the general transformer architecture, the embedding vectors which went through the attention module, fed into the feed forward network that has two linear projection layers and one activation function.
More specifically, the input vector is expanded to a dimension 4.0 times larger than the embedding dimension by a first linear projection layer (as defined by the FFN Expand Ratio in Table \ref{tab:hyperparameters}), then nonlinearly transformed via an GELU\cite{GELU} activation function, and then converted back to a lower dimension by another linear projection layer, where the vector is merged by addition.This process is expressed as follows:
\begin{equation}
  x_{ffn} = GELU(W_1(x))
\end{equation}
\begin{equation}
  x = W_2(x_{ffn}) + x
\end{equation}

\subsection{Data Augmentation for Skeleton Data}\label{sec:data_augmentation}
In deep learning, the amount of data is one of the important aspects in obtaining a better learning representation.
If training is done with insufficient data, the deep learning model tends to overfit.
However, due to high annotation cost, current labeled sign data sources are limited.
To tackle this issue, some works adopted data augmentation~\cite{s23052853} technique, or the self-supervised learning approach~\cite{SignBERT,SignBERT+,zhao2023BEST} in Sign Language Recognition task.

In this experiment, we also adopted some data augmentation techniques to avoid the overfitting.
As basic augmentation technique, we applied random rotation and random shifting to xy skeleton coordinates data.
The rotation range is between -15 degrees and 15 degrees, and the shift ranges from -0.1 to 0.1, with changes applied to the x and y axes separately.
These process is expressed as follows:
\begin{equation}
    \tilde{x} = (x cos\theta - y sin\theta) + \Delta x
\end{equation}
\begin{equation}
    \tilde{y} = (x sin\theta + y cos\theta) + \Delta y
\end{equation}
$\Delta x$ and $\Delta y$ denote as shifting value, and $\theta$ denote as rotation degrees.

In addition, we also adopted the Joint Mixing Data Augmentation~\cite{10.1145/3700878} which can generally improve the effectiveness and robustness.

To handle video inputs of varying lengths and to augment the training data, we employed different frame sampling strategies for the training and inference phases.
For training, we used a random temporal sampling strategy. From a given video sequence $V$ with a total of $T$ frames, we randomly selected a starting frame and extracted a continuous clip of a fixed length $L$. If the video is shorter than the desired length ($T \le L$), we padded the sequence by repeating the last frame until it reaches the length $L$. This acts as a form of data augmentation, ensuring the model sees different segments of the video in each epoch, which improves its robustness.
For validation and testing, we adopted a uniform sampling strategy to ensure a comprehensive and deterministic evaluation. We extracted $k$ evenly-spaced, overlapping clips of length $L$ from the entire video sequence. The starting frame index $s_i$ for the $i$-th clip (where $i \in \{0, 1, \dots, k-1\}$) is determined as:
$$
s_i = \lfloor i \times \frac{T-L}{k-1} \rfloor + 1
$$
The final prediction for the video was obtained by averaging the softmax prediction scores from all $k$ clips. This multi-view inference approach provides a more stable and reliable assessment of the model's performance. In our experiments, we set the clip length $L$ to 50 frames for WLASL dataset, and 32 frames for JSL and KSL, and the number of clips $k$ to 4 for all datasets.

\subsection{Experimental Setting}
The JSL and KSL dataset was split into training, validation, and testing sets using a 8:1:1 ratio, ensuring each gloss had at least one sample in both splits, and keeping user-independent condition that means the each set does not contain data come from the same user.
In other words, each dataset consists of data from different users only, ensuring that no single user's data appears in multiple sets.This condition is crucial for evaluating a model’s generalization performance, as it prevents overfitting to specific users and allows for a more accurate assessment of how the model performs on unseen data.
For WLASL dataset, we evaluated the performance with following the official splitting which is provided by json file.

All experiments were conducted on a machine equipped with a GeForce RTX 5090 GPU (32GB VRAM), 32GB RAM, and CUDA 12.8.1. Our implementation is based on PyTorch \cite{paszke2019pytorch}.

We based our training configuration on a set of standard hyperparameters commonly used in the Transformer and skeleton-based recognition literature, ensuring a robust and reproducible baseline\cite{do2024skateformer}.
Models were trained from scratch using the AdamW optimizer\cite{loshchilov2017decoupled}, a standard choice for Transformer models.
To ensure stable training, we employed a cosine annealing scheduler with a linear warmup of 5 epochs. The base learning rate was set to $1.0 \times 10^{-3}$, a typical and effective starting point for this architecture when paired with AdamW.

The per-device batch size was set to 16, which was the maximum size that could be accommodated by our GPU memory (GeForce RTX 5090, 32GB VRAM). To achieve a larger and more stable effective batch size, we employed gradient accumulation for 16 steps, resulting in an effective batch size of 256.

Unlike large-scale pre-training on datasets like ImageNet, which often converge in fewer epochs, training Transformer models from scratch on small-to-medium-sized datasets like WLASL, JSL, and KSL requires a significantly longer training duration to achieve convergence while avoiding overfitting.
Therefore, we trained models for 1500 epochs on WLASL and 1000 epochs on JSL and KSL. This extended duration was intentionally set to allow the cosine annealing scheduler to fully complete its cycle, ensuring the model could learn robust representations and converge properly from these limited datasets.

For regularization and training stability—common challenges in deep Transformer models, especially on smaller data—we applied a standard set of techniques: label smoothing (0.05), gradient clipping (10.0), FFN dropout (0.25), and Stochastic Depth (0.25).

The detailed model architecture specifications and all training hyperparameters are summarized in Table \ref{tab:hyperparameters}. The output layer's neurons were matched to the number of classes for each dataset (e.g., 100 for WLASL100, 46 for JSL, and 32 for KSL). The specific frame sampling strategies for training and testing are detailed in Section \ref{sec:data_augmentation}.
To ensure a robust evaluation of our model's stability, we did not use a single fixed seed. Instead, we conducted all experiments 10 times, using 10 different random seeds (integers 0 through 9). The results presented in tables (e.g., Table \ref{Tab:Full_Robustness_Results}) report the mean, standard deviation, and best performance across these 10 runs.

\begin{table}[htp]
\centering
\caption{Key hyperparameters for our experiments.}
\label{tab:hyperparameters}
\begin{tabular}{ll}
\toprule
Parameter & Value \\
\midrule
\multicolumn{2}{l}{\textbf{Model Architecture}} \\
\quad Embedding Dimension & 128 \\
\quad Number of Blocks & 10 \\
\quad Number of Heads & 8 \\
\quad Head Dimension & 64 \\
\quad FFN Expand Ratio & 4.0 \\
\midrule
\multicolumn{2}{l}{\textbf{Data \& Training}} \\
\quad Epochs & 1500 \\
\quad Batch Size & 16 \\
\quad Accumulation Steps & 16 \\
\quad (Effective Batch Size) & (256) \\
\quad Input Sequence Length ($L$) & 50 \\
\quad Number of Joints & 55 \\
\quad Test Clips ($k$) & 4 \\
\midrule
\multicolumn{2}{l}{\textbf{Optimization (AdamW)}} \\
\quad Base Learning Rate & 0.001 \\
\quad Weight Decay & 0.01 \\
\quad Betas & (0.9, 0.999) \\
\quad Epsilon & 1.0e-08 \\
\quad LR Scheduler & Cosine \\
\quad Warmup Epochs & 5 \\
\quad Warmup Init LR & 0.0001 \\
\quad Minimum LR & 1.0e-5 \\
\midrule
\multicolumn{2}{l}{\textbf{Regularization}} \\
\quad Label Smoothing & 0.05 \\
\quad Gradient Clip Norm & 10.0 \\
\quad FFN Dropout & 0.25 \\
\quad Stochastic Depth Rate & 0.25 \\
\bottomrule
\end{tabular}
\end{table}

\subsection{Evaluation Metrics}
For evaluation, we reported the classification accuracy following~\cite{li2020word}, \textit{i.e.,} Top-1 (T-1) and Top-5 (T-5) per instance for WLASL dataset, and presented the Top-1 classification accuracy on JSL and KSL dataset.
Top-K accuracy represents the proportion of samples where the correct label appears among the model's top-K predicted classes.
The performance of deep learning models is often sensitive to stochastic elements, including weight initialization and data shuffling order. A single training run may therefore not be representative of the model's true capability. To ensure a fair and robust evaluation and to enhance the reproducibility of our results, we repeated every experiment 10 times, each with a unique random seed. In our result tables, we present the mean accuracy as the primary metric, accompanied by the standard deviation (std) to indicate the stability of the model, and the maximum (max) accuracy to show its peak potential.

\section{Experimental Results}\label{sec5}
In this study, we conducted experiments using sign language and fingerspelling datasets, including JSL, KSL, and WLASL, to evaluate the effectiveness of the proposed model.

\subsection{State of the art similar work comparison for the JSL Dataset} \label{subSec5.4}

\begin{table}[htp]
\centering
\caption{State of the art-similar work comparison of the proposed model with JSL Dataset}
\label{table:jsl_comparison}
% l = left-aligned, c = center-aligned
\begin{tabular}{lllc}
\toprule
Method & Algorithm & Type of Sign & \makecell{Test Best Accuracy\\ (Mean $\pm$ Std) [\%]} \\
\midrule
Ikuno et al.~\cite{IPS_Japan_1} & Random Forest & Static & 79.20 \\
Kwolek et al.~\cite{kwolek2021recognition} & Deep CNN & Static & 92.10 \\
Kobayashi et al.~\cite{wasedathesis_1} & SVM & Static / Dynamic & 65.00 / 96.00 \\
Kakizaki et al.~\cite{s24030826_kakizaki_JSL2} & SVM & Static + Dynamic & 97.20 \\
\midrule
\textbf{Proposed method} & \textbf{Transformer} & \textbf{Static + Dynamic} & \textbf{99.34} (94.80 $\pm$ 2.87) \\
\bottomrule
\end{tabular}
\end{table}

The performance comparison of the proposed model against existing methods on the JSL dataset is presented in Table \ref{table:jsl_comparison}.
Existing methods employ various approaches: Ikuno et al.~\cite{IPS_Japan_1} used Random Forest with hand skeleton features, achieving 79.20\% accuracy; Kwolek et al.~\cite{kwolek2021recognition} utilized a Deep CNN with synthetic data, reaching 92.10\%; and Kobayashi et al.~\cite{wasedathesis_1} applied MSVM to grouped subclasses, achieving 96.00\% for dynamic signs.
As shown in the table, the proposed Transformer-based method achieved a best accuracy of 99.34\% (Mean 94.80 $\pm$ 2.87), recording the highest performance among the compared methods for both static and dynamic signs.

\subsection{Performance Accuracy and State of the Art Comparison of the KSL Dataset} \label{subsubsec5.2.1}

\begin{table}[htp]
\centering
\caption{State-of-the-art comparison for the new KSL dataset.}
\label{tab:new_ksl_sota}
\begin{tabular}{lllc}
\toprule
Method & Modality & Type of Sign & \makecell{Test Best Accuracy\\ (Mean $\pm$ Std) [\%]} \\
\midrule
Shin et al.~\cite{shin2024korean_ksl0} & Skeleton (single frame) & Static & 92.04 \\
\midrule
\textbf{Proposed method} & \textbf{Skeleton (multi-frame)} & \textbf{Static} & \textbf{95.16} (84.60 $\pm$ 10.44) \\
\bottomrule
\end{tabular}
\end{table}

Table \ref{tab:new_ksl_sota} summarizes the performance comparison on the new KSL dataset.
The existing work by Shin et al.~\cite{shin2024korean_ksl0}, which adopted a two-stream feature extraction approach using SVM, reported a classification accuracy of 92.04\%.
In comparison, our proposed model achieved a best test accuracy of 95.16\% (Mean 84.60 $\pm$ 10.44), surpassing the baseline method by approximately 3 percentage points.

\subsection{Performance Accuracy and State of the Art Comparison of the WLASL Dataset} \label{subsubsec5.2.2}

\begin{table}[htp]
\centering
\small
\caption{Detailed performance and robustness evaluation of our model on all subsets of the WLASL dataset (WLASL100, and WLASL300). We report per-instance and per-class Top-1 and Top-5 accuracy (\%) for each of the 10 runs with different random seeds. The mean, standard deviation (std.), and best results are summarized at the bottom.}
\label{Tab:Full_Robustness_Results}
\begin{tabular}{lcccccccc}
\toprule
% --- 1st row header (Dataset names) ---
\multirow{3}{*}{Run (Seed)} & \multicolumn{4}{c}{WLASL100} & \multicolumn{4}{c}{WLASL300} \\
\cmidrule(lr){2-5} \cmidrule(lr){6-9}
% --- 2nd row header (Metric types) ---
 & \multicolumn{2}{c}{Per-Instance (\%)} & \multicolumn{2}{c}{Per-Class (\%)} & \multicolumn{2}{c}{Per-Instance (\%)} & \multicolumn{2}{c}{Per-Class (\%)} \\
\cmidrule(lr){2-3} \cmidrule(lr){4-5} \cmidrule(lr){6-7} \cmidrule(lr){8-9}
% --- 3rd row header (Top-k) ---
 & Top-1 & Top-5 & Top-1 & Top-5 & Top-1 & Top-5 & Top-1 & Top-5 \\
\midrule
Seed 0 & 79.46 & 92.64 & 80.80 & 93.42 & 74.10 & 92.51 & 75.01 & 92.89 \\
Seed 1 & 82.56 & 93.41 & 83.47 & 94.25 & 73.65 & 92.66 & 74.33 & 92.89 \\
Seed 2 & 82.17 & 93.41 & 83.22 & 94.17 & \textbf{74.85} & 92.81 & 75.56 & 93.17 \\
Seed 3 & 82.17 & 94.96 & 83.22 & 95.75 & 73.80 & \textbf{93.11} & 74.45 & \textbf{93.39} \\
Seed 4 & 80.62 & 94.96 & 81.88 & 95.88 & 72.75 & 92.51 & 73.71 & 92.78 \\
Seed 5 & 81.78 & 94.57 & 82.80 & 95.42 & 74.55 & 92.07 & 75.41 & 92.39 \\
Seed 6 & \textbf{82.95} & \textbf{95.35} & \textbf{84.55} & \textbf{95.92} & 73.35 & 92.07 & 74.02 & 92.61 \\
Seed 7 & 80.62 & 93.02 & 81.97 & 94.00 & 73.05 & 92.81 & 73.91 & 93.10 \\
Seed 8 & 80.62 & 93.80 & 81.30 & 94.58 & 74.70 & 91.92 & \textbf{75.63} & 92.39 \\
Seed 9 & \textbf{82.95} & 93.41 & 84.38 & 94.08 & 72.31 & 92.07 & 73.11 & 92.33 \\
\midrule
Mean$\pm$Std. & 81.59 $\pm$ 1.19 & 93.95 $\pm$ 0.93 & 82.76 $\pm$ 1.25 & 94.75 $\pm$ 0.91 & 73.71 $\pm$ 0.86 & 92.45 $\pm$ 0.40 & 74.51 $\pm$ 0.86 & 92.79 $\pm$ 0.36 \\
Best & \textbf{82.95} & \textbf{95.35} & \textbf{84.55} & \textbf{95.92} & \textbf{74.85} & \textbf{93.11} & \textbf{75.63} & \textbf{93.39} \\
\bottomrule
\end{tabular}
\end{table}

\begin{table}[htp]
\centering
\small
\caption{State-of-the-art comparison for skeleton-only methods on the WLASL Dataset. Modality is abbreviated as: S (skeleton), H (hands), P (full keypoints), and RGB (RGB frames). All our results report the best Per-Instance accuracy achieved across 10 runs, with the mean and standard deviation shown in parentheses.}
\label{Table_WLASL_result_com}
\begin{tabular}{llcccccccc}
\toprule
\multirow{2}{*}{Method} & \multirow{2}{*}{Modality} & \multicolumn{2}{c}{WLASL100} & \multicolumn{2}{c}{WLASL300} & \multicolumn{2}{c}{WLASL1000} & \multicolumn{2}{c}{WLASL2000} \\
\cmidrule(lr){3-4} \cmidrule(lr){5-6} \cmidrule(lr){7-8} \cmidrule(lr){9-10}
 & & Top-1 & Top-5 & Top-1 & Top-5 & Top-1 & Top-5 & Top-1 & Top-5 \\
\midrule

Pose-GRU~\cite{li2020word} & S & 46.51 & 76.74 & 33.68 & 64.37 & 30.01 & 58.42 & 22.54 & 49.81 \\
Pose-TGCN~\cite{li2020word} & S & 55.43 & 78.68 & 38.32 & 67.51 & 34.86 & 61.73 & 23.65 & 51.75 \\
GCAR~\cite{10456765_miah_large_scale} & S & 63.25 & 80.00 & 43.80 & 69.01 & --- & --- & 24.10 & 52.62 \\
Sign2Pose\cite{eunice2023sign2pose} & S & 80.9 & --- & 64.21 & --- & 49.46 & --- & 38.65 & --- \\
SignBERT~\cite{SignBERT} & S (H) & 76.36 & 91.09 & 62.72 & 85.18 & --- & --- & 39.40 & 73.35 \\
SignBERT~\cite{SignBERT} & S (H+P) & 79.07 & 93.80 & 70.36 & 88.92 & --- & --- & 47.46 & 83.32 \\
SignBERT+ ~\cite{SignBERT+} & S & 79.84 & 92.64 & 73.20 & 90.42 & --- & --- & 48.85 & 82.48 \\
BEST~\cite{zhao2023BEST} & S & 77.91 & 91.47 & 67.66 & 89.22 & --- & --- & 46.25 & 83.32 \\
SKIM~\cite{SKIM} & S & 74.42 & 90.72 & 73.95 & 92.67 & --- & --- & 55.37 & 86.98 \\
\midrule
Ours (Mean $\pm$ Std.) & S & 81.59 $\pm$ 1.19 & 93.95 $\pm$ 0.93 & 73.71 $\pm$ 0.86 & 92.45 $\pm$ 0.40 & --- & --- & --- & --- \\
\textbf{Ours (Best)} & \textbf{S} & \textbf{82.95} & \textbf{95.35} & \textbf{74.85} & \textbf{93.11} & --- & --- & --- & --- \\
\bottomrule
\end{tabular}
\end{table}

Table \ref{Table_WLASL_result_com} provides a comprehensive comparison of state-of-the-art skeleton-only methods on the WLASL dataset.
Previous skeleton-based methods include RNN-based approaches like Pose-GRU and Pose-TGCN~\cite{li2020word}, and recent Self-Supervised Learning (SSL) approaches such as SignBERT+~\cite{SignBERT+}, which established the previous state-of-the-art with 79.84\% on WLASL100.
Other recent architectures include SKIM~\cite{SKIM}, which utilizes CTR-GCN.

Our proposed model achieved a Top-1 accuracy of 82.95\% on WLASL100 and 74.85\% on WLASL300.
These results represent the highest accuracy among the listed skeleton-only methods, outperforming the previous best result (SignBERT+) on the WLASL100 subset.

\section{Discussion}

This study proposed a Transformer-based model for sign language recognition and evaluated its performance across three datasets: JSL, KSL, and WLASL.
The results demonstrate that the proposed method consistently achieves state-of-the-art performance in skeleton-based recognition, surpassing existing approaches in both fingerspelling (static/dynamic) and large-scale word-level tasks.

Regarding the fingerspelling tasks (JSL and KSL), the proposed model significantly outperformed conventional methods such as Random Forest~\cite{IPS_Japan_1} and SVM~\cite{shin2024korean_ksl0}.
While these traditional classifiers are effective for static poses, they often struggle to capture the temporal transitions inherent in sign language.
The superior performance of our model suggests that the Transformer's self-attention mechanism effectively captures the subtle spatial-temporal dependencies required to distinguish between visually similar signs, such as the dynamic JSL subclasses described by Kobayashi et al.~\cite{wasedathesis_1}.

On the WLASL dataset, a key finding is that our model, trained entirely from scratch, outperformed methods utilizing complex Self-Supervised Learning (SSL) frameworks, such as SignBERT~\cite{SignBERT} and SignBERT+~\cite{SignBERT+}.
Typically, SSL is employed to overcome the lack of labeled data by learning general representations.
However, our results indicate that the proposed sequential spatial-temporal attention mechanism is highly data-efficient, capable of acquiring sufficiently expressive representations directly from the available training data without the need for large-scale pre-training.

We must acknowledge two limitations regarding the scope and comparability of these results.
First, regarding the KSL dataset, the comparison requires nuance.
The KSL dataset consists primarily of static fingerspelling, which Shin et al.~\cite{shin2024korean_ksl0} classified using single-frame data.
Since our model requires a sequence of skeleton data, the input modality differs.
Although our model achieved higher accuracy, the computational cost of processing sequences for purely static targets should be weighed against the efficiency of single-frame classifiers.
Second, as shown in the wider literature (e.g., NLA-SLR), multi-modal methods integrating RGB and skeleton data generally achieve higher absolute accuracy than skeleton-only methods.
Our study was intentionally limited to the skeleton domain to prioritize privacy preservation and robustness to background noise.

Despite these limitations, the establishment of a new state-of-the-art in the skeleton-only domain has significant implications.
It demonstrates that the upper bound of performance for lightweight, privacy-preserving models has not yet been reached.
Future work should examine the integration of this robust Transformer backbone into a multi-modal framework.
Furthermore, given the model's success with training from scratch, applying SSL pre-training to this specific architecture could potentially yield even further performance gains.

\section{Conclusion} \label{sec6}
In this paper, we proposed a novel Stacked Spatial-Temporal Transformer Network for Sign Language Recognition (SLR). Our contributions are summarized as follows:
First, we introduced a hierarchical, stacked architecture where sequential transformers effectively model both short-range and long-range dependencies.
Second, we leveraged dual multi-head attention mechanisms—Spatial MHA (intra-frame) and Temporal MHA (inter-frame)—to capture the complex spatio-temporal dynamics inherent in sign language gestures.
Third, we demonstrated the effectiveness of our approach through extensive experiments, showing that it outperforms existing methods without relying on large-scale pre-training.

We validated our model on diverse datasets, including JSL, KSL, and WLASL.
A key finding is that our model, \textbf{trained entirely from scratch}, achieves state-of-the-art performance in the challenging fingerspelling categories and establishes a new SOTA for skeleton-only methods on WLASL, surpassing several approaches that rely on complex self-supervised pre-training.

Despite these successes, there remains potential for further enhancement.
For future work, we plan to explore two primary directions: 1) adopting a self-supervised learning (SSL) approach to leverage our model's data-efficient architecture for even greater performance, and 2) integrating multi-modal data, such as RGB streams.
Furthermore, we aim to investigate the integration of Large Vision and Language Models to develop next-generation, highly accurate sign language recognition systems.

\section*{Abbreviations}
\begin{table}[htp]
\label{Appendix_A}
\setlength{\tabcolsep}{3pt}
\begin{tabular}{ll}
GCN & Graph Convolutional Network \\
GCAR & Graph convolution with attention and residual connection \\
SLR &Sign Language Recognition\\
ASL& American Sign Language \\
RNN & Recurrent Neural Network \\
LSTM & Long short term memory \\
TGCN&Temporal graph convolutional network\\
ML &Machine learning \\
DL & Deep learning  \\
\end{tabular}
\end{table}

\bibliographystyle{unsrt}
\bibliography{references}

\end{document}